# Teaching Robots to Handle Nuclear Waste: A Teleoperation-Based Learning Approach< – 25429

Joong-Ku Lee *, Hyeonseok Choi *, Young Soo Park **, and Jee-Hwan Ryu *
\* Korea Advanced Institute of Science and Technology
\*\* Argonne National Laboratory

## ABSTRACT

This paper presents a Learning from Teleoperation (LfT) framework that integrates human expertise with robotic precision to enable robots to autonomously perform skills learned from human operators. The proposed framework addresses challenges in nuclear waste handling tasks, which often involve repetitive and meticulous manipulation operations. By capturing operator movements and manipulation forces during teleoperation, the framework utilizes this data to train machine learning models capable of replicating and generalizing human skills. We validate the effectiveness of the LfT framework through its application to a power plug insertion task, selected as a representative scenario that is repetitive yet requires precise trajectory and force control. Experimental results highlight significant improvements in task efficiency, while reducing reliance on continuous operator involvement.

## INTRODUCTION

The handling of nuclear waste poses significant challenges due to the hazardous and complex nature of the tasks involved. These operations often require high precision and careful manipulation in environments where human workers face substantial risks. Traditional teleoperation systems have been widely adopted to improve safety by allowing human operators to control robotic manipulators from a distance. However, these systems rely heavily on continuous human intervention, leading to operator fatigue and limiting efficiency [1]. Furthermore, the dynamic and unpredictable nature of nuclear waste handling environments makes it difficult to achieve full automation using conventional programming approaches.

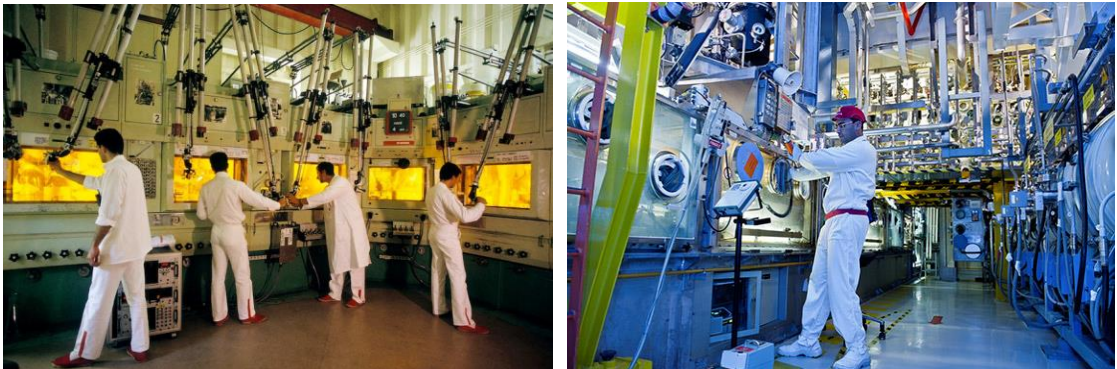

Figure 1. A conventional teleoperation system for nuclear waste handling.

To address these limitations, we propose a Learning from Teleoperation (LfT) framework that leverages human expertise and robotic precision to enable autonomous task execution [2]. This framework bridges the gap between teleoperation and automation by capturing operator movements, manipulation forces, and environmental context during teleoperation sessions. These data are then used to train machine learning models capable of replicating and generalizing human skills. The LfT framework aims to maintain the safety advantages of teleoperation while reducing the need for continuous operator oversight, thereby improving task efficiency and mitigating operator fatigue.

This study focuses on validating the LfT framework in the context of nuclear waste handling tasks, which are often repetitive and meticulous. The framework is applied to power plug insertion task, which is selected as a representative scenario that is repetitive yet requires precise trajectory and force control. Experimental results demonstrate that the proposed approach significantly enhances task efficiency and safety, paving the way for more sustainable and effective robotic solutions in hazardous environments.

## METHODOLOGY

This section introduces the Learning from Teleoperation (LfT) framework, focusing on its design, data acquisition process, and machine learning model for learning human manipulation skills. By capturing operator actions during teleoperation and leveraging this data to train machine learning models, the framework enables robots to replicate and generalize human expertise across diverse task scenarios.

### Overview of the LfT Framework

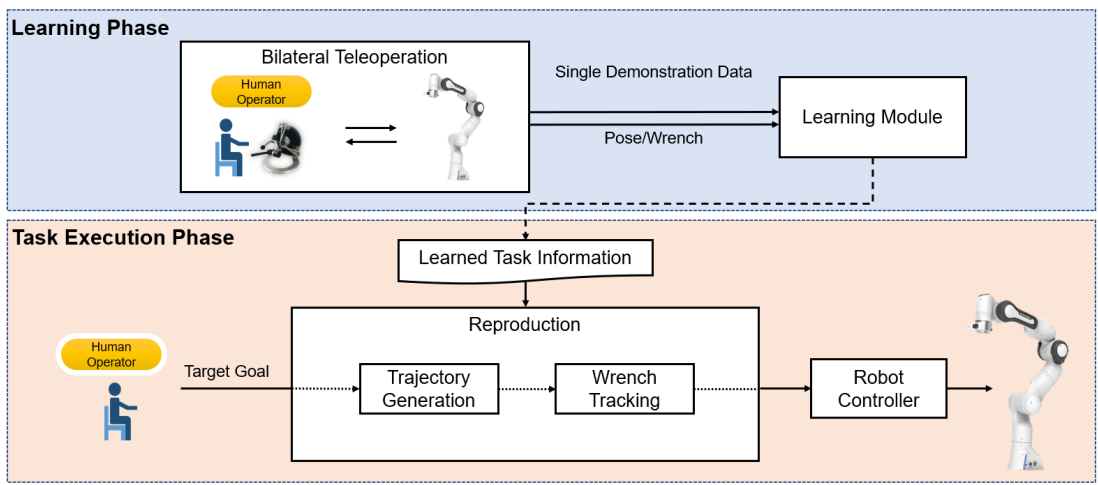

Figure 2. Conceptual diagram of the proposed learning from teleoperation (LfT) framework.

The LfT framework is designed to enable robots to autonomously perform complex manipulation tasks by learning directly from human operators. The framework consists of two main phases: the Learning Phase and the Task Execution Phase, as illustrated in Figure 2.
1. Learning Phase: During this phase, a bilateral teleoperation system is used to capture a demonstration of the task performed by a human operator. Key data, including pose trajectories and wrench (force/torque) information, are collected and processed by the learning module. This phase focuses on extracting and encoding task-specific knowledge into a reusable representation.
2. Task Execution Phase: In the execution phase, the robot leverages the learned task information to autonomously reproduce the demonstrated skill. Given a target goal provided by the operator, the framework generates appropriate motion trajectories and performs wrench tracking to ensure task accuracy. These outputs are processed by the robot controller to execute the task in real-world environments.

This two-phase structure allows the LfT framework to bridge the gap between teleoperation and autonomous robotic operation, ensuring that robots can replicate and generalize human manipulation skills in diverse scenarios.

**Bilateral Teleoperation System**

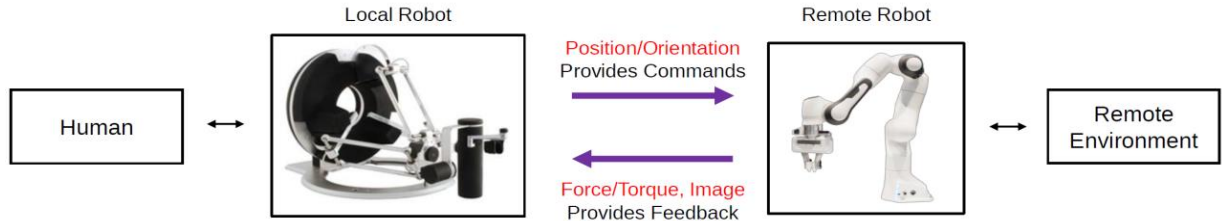

Figure 3. Schematic of the bilateral teleoperation system. The movement of the local robot is transmitted to the remote robot, while force/torque and image feedback are sent back to the operator.

The bilateral teleoperation system serves as the foundational interface for the Learning from Teleoperation (LfT) framework, enabling human operators to intuitively control a remote robot while receiving real-time feedback from the environment. As illustrated in Figure 3, the system facilitates seamless communication between the human operator, the local robot, the remote robot, and the remote environment.

*Human Operator:* The human operator interacts with the local robot, using their expertise to perform tasks in the remote environment. Through this interaction, precise control of the remote robot is achieved.

*Local Robot:* The local robot captures the operator's input in terms of position and orientation, which are translated into commands sent to the remote robot. This device provides a haptic interface that allows the operator to feel forces and torques from the remote environment, creating a realistic and immersive control experience.

*Remote Robot:* The remote robot executes the commands received from the local robot, manipulating objects in the remote environment with precision. Force/torque sensors and visual feedback from cameras enable the remote robot to send real-time feedback to the local robot and human operator.

*Feedback Loop:* The system operates in a bilateral loop where the position/orientation commands are sent from the local robot to the remote robot, and force/torque and visual feedback are transmitted back to the local robot. This closed-loop feedback ensures precise task execution and allows the operator to adjust actions dynamically.

**Data Acquisition**

The data acquisition process is a fundamental step in the Learning from Teleoperation (LfT) framework, facilitating the collection of high-quality demonstration data during teleoperation. During this phase, the system captures detailed information about the operator's actions, which forms the basis for training the learning module. Specifically, the system records the position and orientation of the robotic end-effector in Cartesian space, capturing precise trajectories necessary for accurate task reproduction. Alongside this, interaction forces and torques exerted by the robotic manipulator during task execution are also collected, which is particularly critical for tasks requiring delicate manipulation or physical interaction with objects.

To ensure consistency and accuracy, all data streams, including position, orientation, force, and torque information, are time-synchronized before storage. This synchronization aligns the temporal relationship between these variables, which is essential for maintaining the integrity of the data during learning. The collected data are stored in a hierarchical format using the HDF5 file structure, enabling efficient storage and retrieval of large-scale datasets.

**Machine Learning Model**

The LfT framework utilizes a combination of machine learning techniques to effectively encode and reproduce human manipulation skills. Two distinct approaches are employed to learn and generalize the positional and force/torque data captured during the teleoperation phase: Dynamic Movement Primitives (DMPs) for learning position trajectories and Gaussian Mixture Models (GMMs) combined with Gaussian Mixture Regression (GMR) for encoding and inferring force/torque behavior.

*Learning Position Trajectories with DMPs:* Dynamic Movement Primitives (DMPs) are a widely used framework for learning and generalizing motion trajectories [3,4]. In the LfT framework, DMPs are employed to model the end-effector's positional and orientational movements during task demonstrations. DMPs encode the demonstrated trajectories into a set of parameters that represent time-dependent motion patterns. By leveraging differential equations, DMPs ensure smooth and adaptable trajectory generation, enabling the robot to generalize the learned motion to new start and goal positions. This flexibility is crucial for tasks with varying spatial configurations, allowing the robot to maintain task accuracy even in dynamic environments.

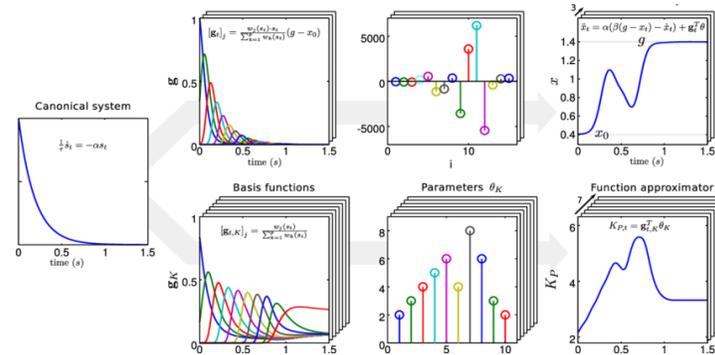

Figure 4. Dynamic Movement Primitives (DMPs). The core idea behind DMPs is to model movements to simple linear dynamical system with a non-linear perturbation component. The trajectory information is encoded using DMPs.

*Encoding and Inferring Force/Torque Behavior with GMM and GMR:* For modeling interaction forces and torques, the LfT framework leverages Gaussian Mixture Models (GMMs) to encode the multi-dimensional force/torque data [5]. The advantages of GMMs include their capacity to represent complex, non-linear distributions of data through a mixture of Gaussian components. Once the force/torque data is encoded, Gaussian Mixture Regression (GMR) is used to infer force/torque values required for task execution [6]. GMR provides smooth, continuous predictions of force/torque values based on contextual variables, such as time or position, allowing the robot to adapt its behavior in real-time. This is particularly useful for tasks requiring precise force control, such as manipulating delicate objects.

The positional data learned through DMPs and the force/torque behavior encoded with GMM/GMR are integrated during task execution. This ensures that the robot's movements are not only accurate but also physically responsive to environmental interactions. By combining these two complementary approaches, the LfT framework achieves robust and adaptable skill reproduction across diverse task scenarios.

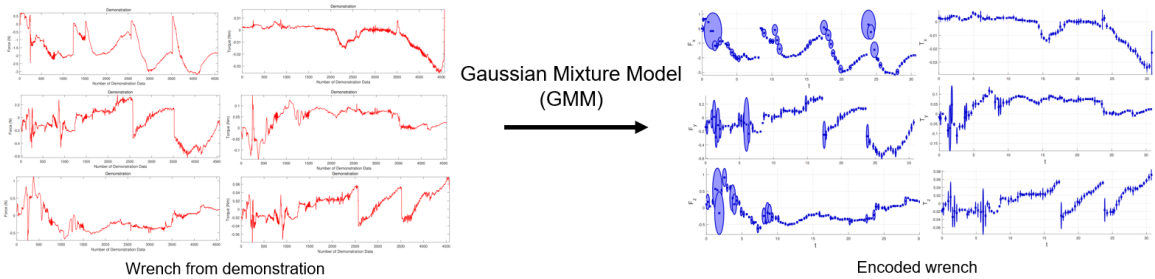

Figure 5. Demonstrated force/torque encoding results with GMM.

### EXPERIMENTAL VALIDATION

The experimental validation of the LfT framework focuses on assessing its effectiveness in learning and replicating human manipulation skills across a range of nuclear waste handling tasks. The experiments are designed to evaluate the system's ability to perform tasks autonomously after being trained on a single teleoperated demonstration. This section provides an overview of the teleoperation setup, the target tasks, and the test results.

**Experimental Setup**

A bilateral teleoperation system is employed as the interface for data collection during the learning phase. As illustrated in Figure 6, this setup consists of a local robot, operated by a human user, and a remote robot that executes tasks in a remote environment.

The local system includes a bimanual haptic display, which is implemented using a collaborative robotic arm, a VR headset, and a haptic glove. The robot arm-based bimanual haptic display offers comprehensive coverage of the human operator's workspace and provides sufficient force feedback. The VR headset delivers a real-time immersive view of the remote scene [7], while the haptic glove, integrated with the bimanual haptic display, captures the operator's grasping input by measuring flexion.

On the remote side, there are two types of remote robots:
1. Bimanual remote robot – This robot follows the operator's arm movements and is equipped with force/torque sensors to capture and reflect interaction data. It also has a gripper system to replicate the operator's grasp and manipulate objects accordingly.
2. Head-tracking remote robot – This robot functions as the "head" of the remote system. It tracks the operator's head pose and the camera system updates the remote view in real time, ensuring that head movements made by the operator are accurately mirrored in the remote environment.

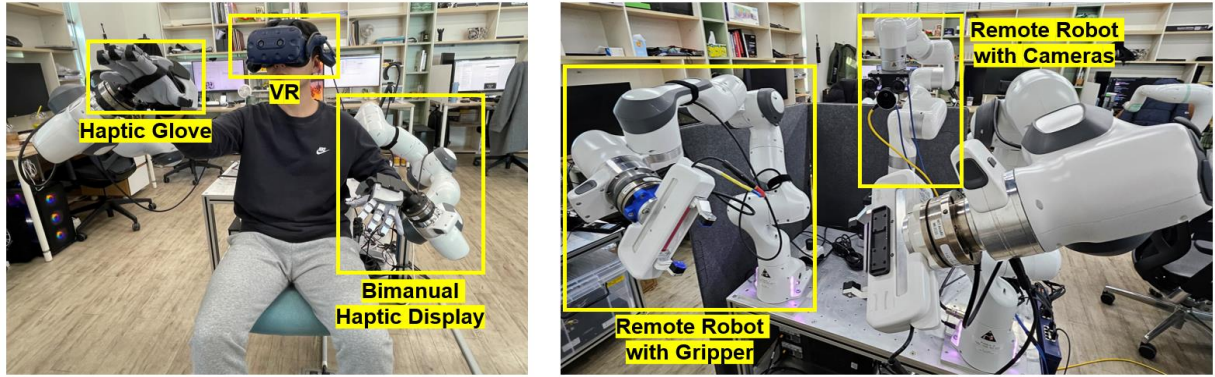

Figure 6. (Left) The local human interface, consisting of a robot arm-based bimanual haptic display, a haptic glove, and a VR headset. (Right) The remote robot system, including a bimanual robot equipped with a gripper to follow the operator's arm movements, and a camera-equipped robot to track the operator's head movements.

**Target Task**

In this study, we adopt power plug insertion as the primary task to validate our Learning from Teleoperation (LfT) framework. Although seemingly straightforward, it is repetitive yet demands precise trajectory planning and stable force control, mirroring the challenges of hazardous environments like nuclear waste management. By selecting a task that requires frequent repetition, we can collect rich teleoperation data to refine the LfT-based learning process, ultimately enhancing task efficiency and reducing reliance on continuous operator involvement. Consequently, power plug insertion effectively represents a broader class of repetitive, high-precision operations in waste handling, thereby demonstrating the scalability and robustness of our proposed approach.

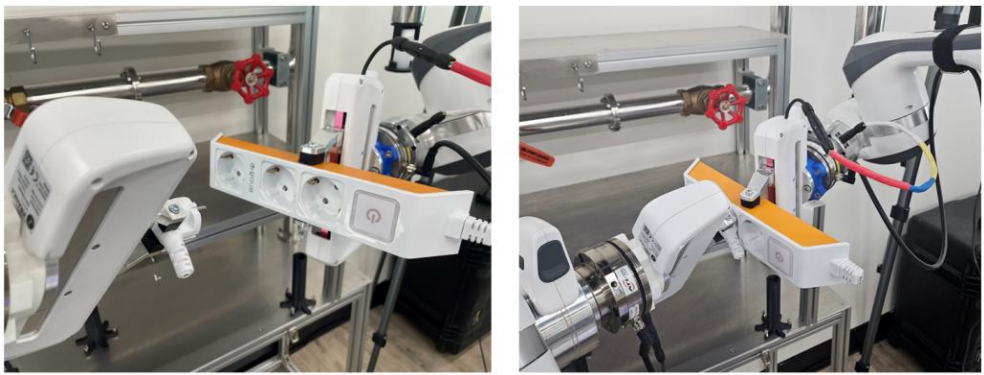

Figure 7. The target task is power plug insertion, representing a repetitive, high-precision manipulation scenario commonly encountered in waste handling.

**Test Results**

As illustrated in Figure 8, the proposed LfT framework is validated in a power plug insertion experiment, where an initial demonstration of the task provides the training data. The framework encodes and learns trajectory information using Dynamic Movement Primitives (DMP), while force/torque interactions are captured via Gaussian Mixture Models (GMM). To assess its reproduction performance under varied conditions, the initial pose of the robot arm is deliberately altered, simulating the real-world scenarios of repetitive manipulation with different starting configurations.

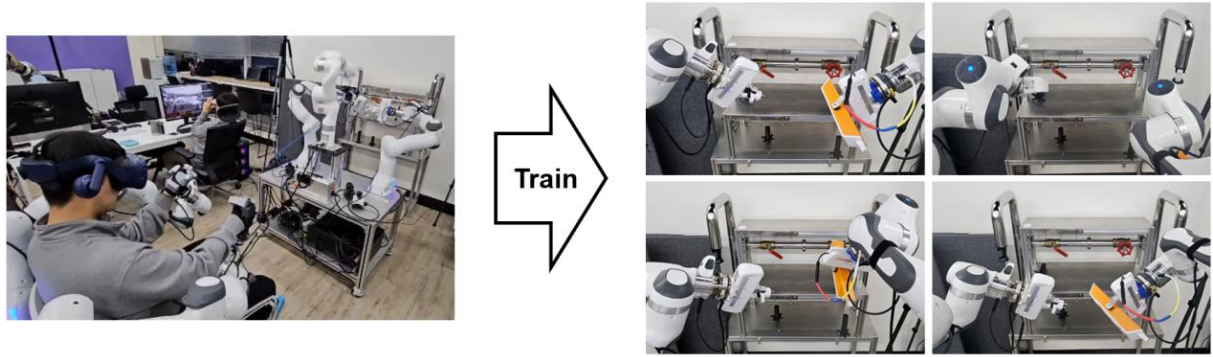

Figure 8. The proposed LfT framework utilizes an initial demonstration to train both trajectory and force/torque interactions, then reproduces the trained task under various conditions.

Table 1 summarizes the results of verifying the reproduction success of the agent—trained under the proposed LfT framework—across ten different initial conditions. Among these ten conditions, the agent failed only once, resulting in a 90% success rate. The single failure occurred when the plug and the power strip collided during the approach phase, as no avoidance algorithm had been implemented for the gripped object. However, this incident arose under a highly extreme variation in the initial condition and is unlikely to occur in typical scenarios.

Table 1. Success or failure of the power plug insertion task under various initial conditions.

| Initial Condition | Result[A] | Initial Condition | Result |
|---|---|---|---|
|  | O |  | O |
| Initial Condition | Result | Initial Condition | Result |
|  | O |  | X |
| Initial Condition | Result | Initial Condition | Result |

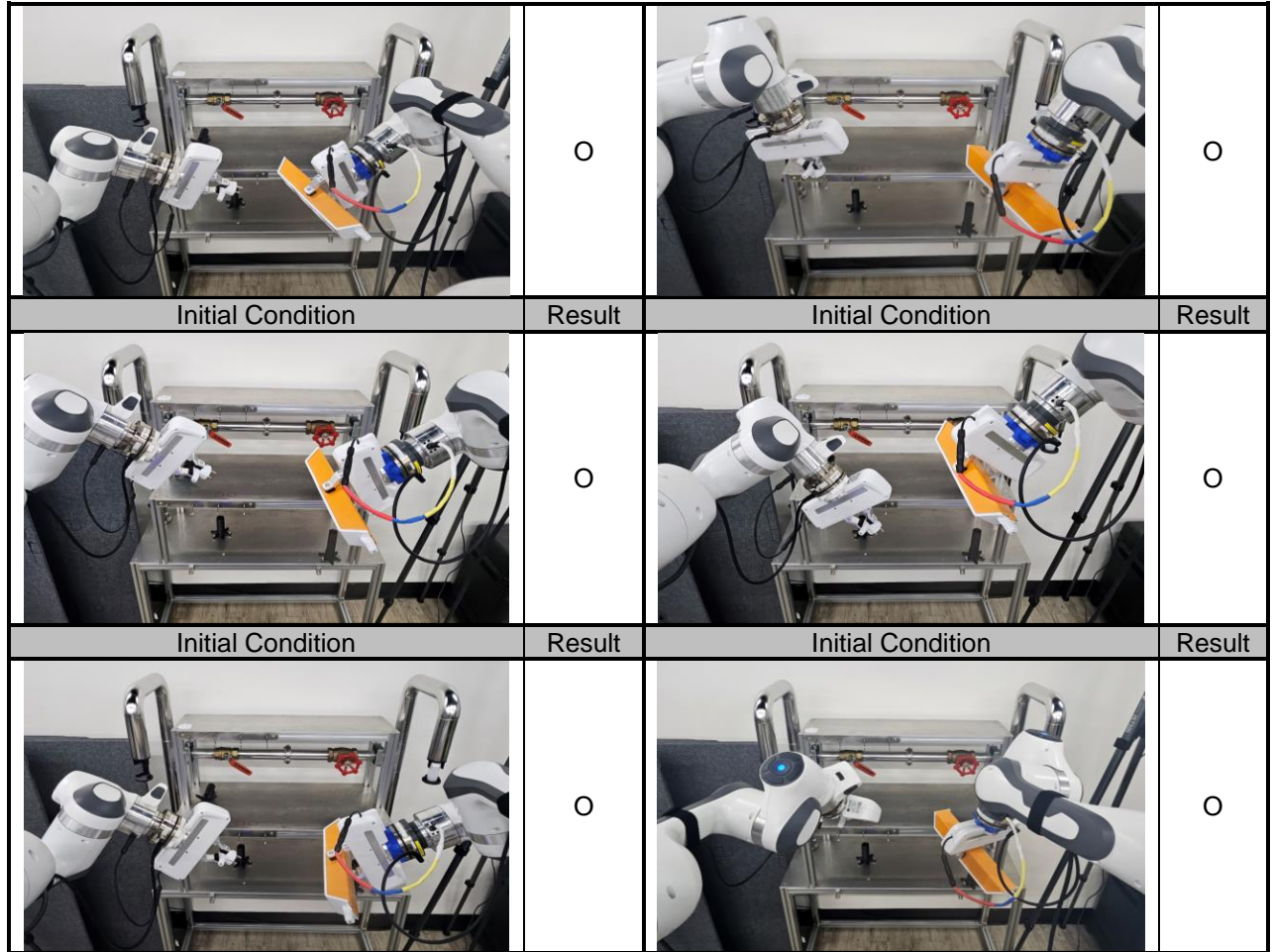

[A] Task outcome from the initial condition. O = Success, X = Fail.

Figure 9 provides a visualization of how the force/torque data from the demonstration is reproduced during the task. As illustrated in the plot, the system exerts an amount of force during the insertion process that closely matches the forces experienced in the demonstration, indicating that precise force application in reproduction directly contributes to successful task completion. This ability to follow the demonstrated force/torque profile highlights the suitability of the proposed framework not only for insertion tasks but also for other operations requiring fine force control, such as grinding or sanding.

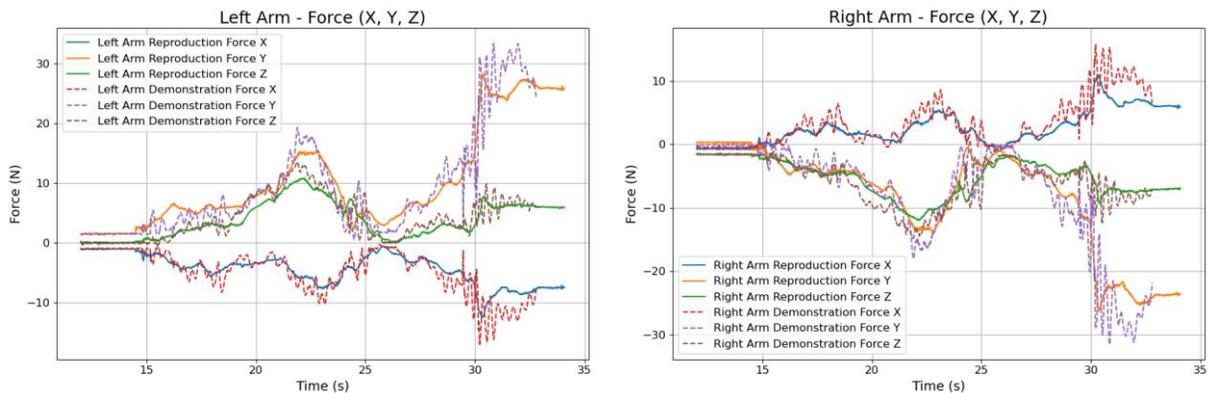

Figure 9. Interaction force/torque data from both the demonstration and the reproduction. The robot aims to replicate the demonstrated force/torque profile during the reproduction phase.

## SUMMARY AND CONCLUSIONS

This paper presented the Learning from Teleoperation (LfT) framework, which enables robots to autonomously learn and perform complex manipulation tasks by leveraging human expertise captured during teleoperation. The framework combines Dynamic Movement Primitives (DMPs) for learning position trajectories and Gaussian Mixture Models (GMMs) with Gaussian Mixture Regression (GMR) for encoding and reproducing interaction forces/torques. Through this approach, the LfT framework bridges the gap between teleoperation and automation, allowing robots to replicate and generalize human manipulation skills.

Experimental validation demonstrated the framework's effectiveness in executing power plug insertion task, which is selected as a representative task that is repetitive and requires precise trajectory and force control. The results showed high task success rates, robustness to the various conditions, and accurate force reproduction, highlighting the framework's ability to perform tasks autonomously with precision and adaptability.

In conclusion, the LfT framework represents a significant step forward in enabling robots to learn and autonomously execute complex tasks in hazardous environments. By combining human expertise with robotic precision, the framework holds great potential for applications in nuclear waste management, industrial automation, and beyond, paving the way for safer, more efficient, and autonomous robotic systems.

## ACKNOWLEDGEMENT

This work was performed in collaboration with Argonne National Laboratory with the supported from the U.S. Department of Energy, Office of Environment Management, Office of Technology Development.